\documentclass[10pt,twocolumn,letterpaper]{article}

\usepackage[pagenumbers]{cvpr} 

\usepackage{graphicx}
\usepackage{amsmath}
\usepackage{amssymb}
\usepackage{booktabs}
\usepackage{microtype}
\usepackage[pagebackref,breaklinks,colorlinks]{hyperref}

\usepackage[capitalize]{cleveref}
\crefname{section}{Sec.}{Secs.}
\Crefname{section}{Section}{Sections}
\Crefname{table}{Table}{Tables}
\crefname{table}{Tab.}{Tabs.}

\usepackage[dvipsnames]{xcolor}

\usepackage{overpic}
\usepackage{enumitem} 
\usepackage{overpic} 
\usepackage{color}
\usepackage{mathtools} 
\usepackage{currfile}
\usepackage{multirow}

\definecolor{turquoise}{cmyk}{0.65,0,0.1,0.3}
\definecolor{purple}{rgb}{0.65,0,0.65}
\definecolor{dark_green}{rgb}{0, 0.5, 0}
\definecolor{orange}{rgb}{0.8, 0.6, 0.2}
\definecolor{red}{rgb}{0.8, 0.2, 0.2}
\definecolor{darkred}{rgb}{0.6, 0.1, 0.05}
\definecolor{blueish}{rgb}{0.0, 0.3, .6}
\definecolor{light_gray}{rgb}{0.7, 0.7, .7}
\definecolor{pink}{rgb}{1, 0, 1}
\definecolor{greyblue}{rgb}{0.25, 0.25, 1}

\newcommand{\CIRCLE}[1]{\raisebox{.5pt}{\footnotesize \textcircled{\raisebox{-.6pt}{#1}}}}

\DeclareMathOperator*{\argmin}{arg\,min}

\newcommand{\loss}[1]{\mathcal{L}_\text{#1}}
\newcommand{\expect}{\mathbb{E}}
\newcommand{\real}{\mathbb{R}}

\newcommand{\eq}[1]{(\ref{eq:#1})}

\newcommand{\Sec}[1]{Sec.~\ref{sec:#1}}

\renewcommand{\paragraph}[1]{\vspace{.5em}\noindent\textbf{#1}.}

\usepackage{lipsum}

\usepackage{blindtext}

\newcommand{\image}{I}
\newcommand{\C}{\boldsymbol{C}} 
\newcommand{\gt}{\text{gt}} 
\newcommand{\exposure}{\Gamma}
\newcommand{\params}{\boldsymbol{\theta}}
\newcommand{\near}{{t_n}}
\newcommand{\far}{{t_f}}
\newcommand{\radiance}{\mathbf{c}}
\newcommand{\sky}{\text{sky}}
\newcommand{\ray}{\mathbf{r}}
\newcommand{\lidarPoint}{\mathbf{p}}
\newcommand{\origin}{\mathbf{o}}
\newcommand{\dir}{\mathbf{d}}
\newcommand{\latentExposure}{\beta}
\newcommand{\density}{\sigma}
\newcommand{\feature}{\mathbf{z}}
\newcommand{\definedAs}{=}
\newcommand{\lidarPoints}{\mathcal{D}}

\newcommand{\gaussian}{\mathcal{N}}

\newcommand{\lossSky}{\loss{seg}}
\newcommand{\lossNear}{\loss{near}}
\newcommand{\lossDepth}{\loss{depth}}
\newcommand{\lossEmpty}{\loss{empty}}

\newcommand{\tickbox}{{\makebox[0pt][l]{$\square$}\raisebox{.15ex}{\hspace{0.1em}$\checkmark$}}}
\newcommand{\untickbox}{{\makebox[0pt][l]{$\square$}}}

\newcommand{\vspacebeforecaption}{-1em}
\newcommand{\vspaceaftercaption}{-1em}
\newcommand{\vspacebeforeTABLEcaption}{-1em}

\newcommand{\kostas}[1]{{\color{Bittersweet} {[\bf Kosta: #1]}}}
\newcommand{\andrew}[1]{{\color{BlueViolet} {[Andrew: #1]}}}
\newcommand{\vitto}[1]{{\color{OrangeRed} {[Vitto: #1]}}}
\newcommand{\JB}[1]{{\color{OliveGreen} {[Jon: #1]}}}
\newcommand{\tom}[1]{{\color{RoyalPurple} {[Tom: #1]}}}
\newcommand{\pratul}[1]{{\color{Emerald} {[Pratul: #1]}}}
\newcommand{\at}[1]{{\color{blueish}#1}}
\newcommand{\AT}[1]{{\color{blueish}{\bf [Andrea: #1]}}}
\newcommand{\At}[1]{\marginpar{\tiny{\textcolor{blueish}{#1}}}}
\newcommand{\al}[1]{\textbf{\color{orange}[AL: #1]}}
\renewcommand{\kostas}[1]{}
\renewcommand{\andrew}[1]{}
\renewcommand{\vitto}[1]{}
\renewcommand{\JB}[1]{}
\renewcommand{\tom}[1]{}
\renewcommand{\pratul}[1]{}
\renewcommand{\at}[1]{}
\renewcommand{\AT}[1]{}
\renewcommand{\At}[1]{}
\renewcommand{\al}[1]{}

\newcommand{\norm}[1]{\left\lVert#1\right\rVert}

\newcommand{\supplementary}{{\color{purple}supplementary material}\xspace}

\usepackage{tabularx}
\usepackage{multirow}
\usepackage{arydshln}

\def\cvprPaperID{2829} 
\def\confName{CVPR}
\def\confYear{2022}

\begin{document}
\title{Urban Radiance Fields}
\author{
Konstantinos Rematas\textsuperscript{1}
\quad
Andrew Liu\textsuperscript{1}
\quad
Pratul Srinivasan\textsuperscript{1}
\quad
Jonathan Barron\textsuperscript{1}
\\
Andrea Tagliasacchi\textsuperscript{1,2}
\quad
Thomas Funkhouser\textsuperscript{1}
\quad
Vittorio Ferrari\textsuperscript{1}
\\[1em]
\textsuperscript{1} Google Research
\quad
\textsuperscript{2}{University of Toronto}
}

\maketitle
\begin{abstract}
    The goal of this work is to perform 3D reconstruction and novel view synthesis from data captured by scanning platforms commonly deployed for world mapping in urban outdoor environments (e.g., Street View).
Given a sequence of posed RGB images and lidar sweeps acquired by cameras and scanners moving through an outdoor scene, we produce a model from which 3D surfaces can be extracted and novel RGB images can be synthesized.
Our approach extends Neural Radiance Fields, which has been demonstrated to synthesize realistic novel images for small scenes in controlled settings, with new methods
for leveraging asynchronously captured lidar data,
for addressing exposure variation between captured images,
and for leveraging predicted image segmentations to supervise densities on rays pointing at the sky.
Each of these three extensions provides significant performance improvements in experiments on Street View data.
Our system produces state-of-the-art 3D surface reconstructions and synthesizes higher quality novel views in comparison to both traditional methods (e.g.~COLMAP) and recent neural representations (e.g.~Mip-NeRF).

\end{abstract}
\section{Introduction}
\label{sec:intro}


In this work we investigate neural scene representations for world mapping, with the goal of performing 3D reconstruction and novel view synthesis
from data commonly captured by mapping platforms such as Street View~\cite{GoogleStreetView}.
This setting features large outdoor scenes, with many buildings and other objects, natural illumination from the sun, and is generally less controlled than previous work~\cite{Mildenhall20eccv_nerf, mildenhall2019llff}.
We focus on street-level mapping: a person carrying a camera rig with a lidar sensor placed on a backpack walking through a city. The camera captures panoramas of the street scene while the lidar sensor reconstructs a 3D point cloud.


Street-level mapping is challenging for neural representations, as the area of interest covers a large area, usually hundreds of square meters. This significantly differs from previous works, which largely focus on either synthetic data~\cite{sitzmann2019srns, Park_2019_CVPR} or small regions of real scenes~\cite{Mildenhall20eccv_nerf, mildenhall2019llff, Yu21cvpr_pixelNeRF, Trevithick21iccv_GRF, Chibane_SRF, Chen21iccv_MVSNeRF}.
Moreover, the scenes contain a large variety of objects, both in terms of geometry and appearance (e.g. buildings, cars, signs, trees, vegetation).
The camera locations are biased towards walking patterns (e.g. walking a straight line) without focusing on any particular part of the scene. This results in parts of the scene being observed by only a small number of cameras, in contrast to other datasets~\cite{Knapitsch2017, mildenhall2019llff, phototourism, reizenstein21co3d} which capture scenes uniformly with a large number of cameras.
Furthermore, the sky is visible in most street scenes, introducing an infinitely distant element that behaves differently than the solid structures near the cameras.
The images typically have highly varying exposures as the cameras use auto-exposure, and the illumination brightness varies depending on the sun's visibility and position.
Combined with auto white balance, this results in the same structure having different colors when observed from different cameras.
Finally, the lidar points have lower resolution in distant parts of the scene, and are even completely absent in some parts of the scene (e.g., for shiny or transparent surfaces).

\begin{figure}[t]
\begin{center}
\includegraphics[width=\linewidth]{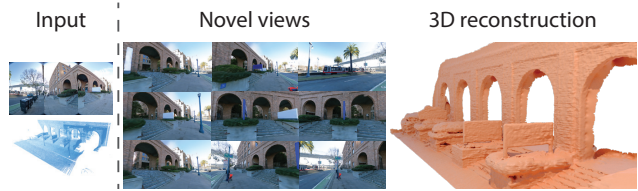}
\end{center}
\vspace{\vspacebeforecaption}
\caption{
\textbf{Overview -- }
Given a set of panoramas and lidar observations from an urban setting, we estimate a neural representation that can be used for novel view synthesis and accurate 3D reconstruction.
}
\label{fig:\currfilebase}
\vspace{\vspaceaftercaption}
\end{figure}


In this paper we extend the popular NeRF~\cite{Mildenhall20eccv_nerf} model in three ways to tailor it to the unique features of the Street View setting and to tackle the challenges above.
First, we incorporate lidar information in addition to RGB signals. By carefully fusing these two modalities, we can compensate for the sparsity of viewpoints in such large scale and complex scenes. We introduce a series of lidar-based losses that allow accurate surface estimation both for solid structures like buildings and for volumetric formations such as trees/vegetation.
Second, we automatically segment sky pixels and define a separate dome-like structure to provide a well-defined supervision signal for camera rays pointing at the sky.
Third, our model automatically compensates for varying exposure by estimating an affine color transformation for each camera.

During experiments with real world data from Street View~\cite{GoogleStreetView}, we find that these three NeRF extensions significantly improve over the state-of-the-art both in the quality of synthesized novel views (+19\% PSNR over \cite{MartinBrualla21cvpr_nerfw}) and 3D surface reconstructions (+0.35 F-score over \cite{kangle2021dsnerf}).
We encourage the reader to view the supplementary material for more results and animated visualizations.

\section{Related Works}
\label{sec:related}
\paragraph{Novel View Synthesis} The 3D reconstruction of urban environments has been studied for decades \cite{musialski2013survey}.  
Most prior work represents the geometry of a city with raw point clouds, acquired either from structure-from-motion \cite{agarwal2011building} or lidar sensors \cite{ilci2020high}, and provide only a sampled, partial representation from which it is difficult to render high quality novel views.  Traditional surface reconstruction methods aggregate the raw data into explicit 3D scene representations, such as textured meshes \cite{romanoni2017mesh} or primitive shapes \cite{debevec1996modeling}.  These methods generally utilize hand-crafted reconstruction algorithms that work best for scenes with large, diffuse surfaces --- a property that does not hold for most urban environments.
Others reconstruct volumetric representations, such as voxels \cite{shim20113d}, octrees \cite{truong2014octree}, or multi-plane images \cite{srinivasan2019pushing,Zhou18siggraph_mpi,flynn2016deepstereo}.  However, these approaches usually have limited resolution or suffer due to the large storage requirements of discretized volumes. 

\paragraph{NeRF} Neural Radiance Fields represent a scene with a multilayer perceptron (MLP) that maps a 3D position and direction to a density and radiance that can be used to synthesize arbitrary novel views with volumetric rendering~\cite{Mildenhall20eccv_nerf}.  Typically this representation is trained per scene with a loss measuring photometric consistency with respect to a collection of posed RGB images.  If the input images are dense  and diverse enough, the scene is small enough, the camera poses are accurate enough, the camera exposure parameters are constant, and the scene is static, the original NeRF model can synthesize remarkably detailed and accurate novel views.

\paragraph{NeRF in vitro}
Many researchers have investigated extensions to NeRF to overcome some of its limitations \cite{Dellaert2021Arxiv}.
Mip-NeRF~\cite{barron2021mipnerf} proposed a scale-aware scene representation based on conical frustrums instead of rays to compensate for blurring and aliasing artifacts.
NSVF~\cite{Liu20neurips_sparse_nerf} reduced the rendering time of radiance fields using an octree representation. Another line of research focused on radiance field estimation from single images~\cite{Yu21cvpr_pixelNeRF, Jang21iccv_CodeNeRF, Rematas21icml_ShaRF}. Radiance fields are also used for surface extraction, either using an SDF representation~\cite{Yariv20neurips_MVNeuralSurfaceRecon, wang2021neurips_neus, Yariv21arxiv_volumenis} or solid surfaces~\cite{Oechsle21ICCV_unisurf}.
However, most of this work has been demonstrated only for input images that are synthetic or captured in a laboratory setting~(``in vitro'')~with complete control over lighting, viewpoint, and scene composition,  and thus cannot be used directly in real-world applications where sensors move along constrained trajectories.

\paragraph{NeRF in situ}
Some research has been directed towards using NeRF-like models for 3D reconstruction and novel view synthesis from images captured in natural environments~(``in situ'')~\cite{Gao-freeviewvideo, Li21iccv_MINE, Sucar21iccv_iMAP, Yang21iccv_ObjectNeRF, Zhi21iccv_SemanticNeRF}. NeRF++~\cite{kaizhang2020_nerfplusplus} investigates the parameterization of unbounded scenes.
Other works have addressed short video inputs \cite{Gao-freeviewvideo} or monocular input \cite{Li21iccv_MINE}. 
Methods like~\cite{kangle2021dsnerf,Wei21iccv_NerfingMVS} use the 3D points from SfM to guide the training of the radiance field.
In work more similar to ours, IMAP \cite{Sucar21iccv_iMAP} performs real-time SLAM from RGB-D images captured with a hand-held camera moving through indoor scenes from the Replica Dataset \cite{replica19arxiv}. 
The work of Azinović \etal~\cite{neuralrgbd} performs surface reconstruction of indoor environments, also using images from an RGB-D camera.
ObjectNeRF \cite{Yang21iccv_ObjectNeRF} and SemanticNeRF \cite{Zhi21iccv_SemanticNeRF} perform novel view synthesis and semantic segmentation from RGB-D videos of the ScanNet dataset \cite{dai2017scannet}.  These systems have been demonstrated for RGB-D data captured in \textit{indoor} scenes, which do not exhibit most of the issues we address. 

\paragraph{NeRF for world mapping}
There has been very little work on using NeRF in outdoor mapping applications.
Neural Scene Graphs \cite{Ost21cvpr_NSG} consider novel view synthesis from images provided with the KITTI Dataset \cite{Geiger2012CVPR}, and NeRF in the Wild (NeRF-W) \cite{MartinBrualla21cvpr_nerfw} does the same for internet photo collections.
Neither system leverages lidar data, which is available in most outdoor mapping platforms \cite{caesar2019nuscenes, chang2019argoverse, geiger2013vision, geyer2020a2d2, huang2018apolloscape, Liao2021Kitti360, lyft2019, sun2019scalability, tong2020cspc} nor do they attempt to extract 3D surface reconstructions.
In addition to this new functionality, in comparison to NeRF-W we also provide improved methods for handling exposure variations and the challenge posed by the sky.

\subsection{Review of Neural Radiance Fields}
\label{sec:nerf}
Neural radiance fields fit a coordinate-based neural network with parameters $\params$ to describe a volumetric scene from a set of posed images $\{\image_i\}_{i=1}^N$; i.e.~with known intrinsic and extrinsic calibration.
To render an image, NeRF uses ray marching to sample the volumetric radiance field and composites the sampled density and color to render the incoming radiance of a particular ray, and supervises the training of $\params$ by an L2 photometric reconstruction loss:
\begin{equation}
\loss{rgb}(\params) = \sum_{i} \expect_{\ray \sim \image_i}
\left[
\norm{\C(\ray) - \C^\gt_i(\ray)}_2^2 
\right]
\label{eq:nerfrgb}
\end{equation}
where $\C^\gt_i(\ray)$ is the ground truth color of ray $\ray$ passing through a pixel in image $i$, and the color $\C(\ray)$ is computed by integrating the weighted volumetric radiance within the ray's near and far bounds $\near$ and $\far$:
\begin{align}
\C(\ray) = \int_\near^\far \!\!\!
w(t)
\cdot
\!\!\underbrace{\radiance(t)}_\text{radiance} \!\! \, dt
\label{eq:nerfcolor}
\end{align}
and $\ray(t)=\origin + t\mathbf{d}$ represents a ray with camera origin $\origin$ oriented as $\mathbf{d}$, with volume rendering integration weights:
\begin{align}
w(t) = \underbrace{\exp \left( -\int_{t_n}^{t} \density(s) \, ds \right)}_\text{visibility of $\ray(t)$ from $\origin$} \:\: \cdot \!\!\!\! \underbrace{\density(t)   \vphantom{\int_{t_n}^{t}}}_\text{density at $\ray(t)$}
\label{eq:weights}
\end{align}
while the intermediate features $\feature(t)$, the volumetric density $\density(t)$ and view-dependent radiance fields $\radiance(t)$ are stored within the parameters $\params$ of fully connected neural networks:
\begin{align}
\feature(t) &= \feature(\ray(t); \params) : \real^3 \rightarrow \real^z 
\label{eq:feature}
\\
\density(t) &= \density(\feature(t); \params) : \real^z \rightarrow \real^+ 
\label{eq:density}
\\
\radiance(t) &= \radiance(\feature(t), \dir; \params) : \real^z \times \real^3 \rightarrow \real^3
\label{eq:radiance}
\end{align}
The discretization of the integrals in \eq{nerfcolor} and \eq{weights} follows~\cite{nerf}. As the scenes used in this work are observed at different distances, we use the integrated positional encoding proposed in mip-NeRF~\cite{barron2021mipnerf} for 3D points $\ray(t)$ and the original positional encoding~\cite{Mildenhall20eccv_nerf} for the viewing direction $\mathbf{d}$.

\section{Data}
\label{sec:data}

This paper investigates how to reconstruct 3D surfaces and synthesize novel views of urban spaces from data commonly collected for autonomous driving and world mapping applications.  Though many suitable data sources are available, we focus our experiments on Trekker data from Street View \cite{GoogleStreetView}, which was acquired from Google with permission via personal communication. 

Street View data is particularly interesting because it has been captured for large parts of the world, and thus provides opportunities for visualization and geometry analysis applications at scale.   However, Street View differs from other 3D scene reconstruction datasets such as Phototourism~\cite{phototourism} in several crucial ways. First, the number of images captured for a particular scene is significantly smaller than those found for popular landmarks. This results in limited diversity of viewpoints. Second these panoramic captures are often accompanied by lidar sensors which provide accurate, but sparse, depth information. 

\paragraph{Image Data}
Street View imagery is collected by multiple fisheye cameras attached to a~\textit{trekker} capturing rig.  Each camera is calibrated with estimated intrinsic parameters and poses relative to the trekker. Images are collected from each camera at approximately 2Hz as the trekker moves through the world.   Images are posed automatically within a global coordinate system using structure-from-motion and GPS information, allowing us to assemble camera rays with origin $\origin$ and direction $\dir$ corresponding to each pixel.

Real world urban scenes have moving objects whose positions change as images are captured over time~(pedestrians, cars, bicyclists, etc).
If unaddressed, these objects can result in trained NeRFs that produce ghosting and blurring.
Because dynamics are often tied to semantics, we run a pre-trained semantic segmentation model~\cite{deeplabv3plus2018} on every image, and then mask pixels of people, which are the most prominent moving category.
 
\paragraph{Lidar Data}
In addition to imaging sensors, the trekker contains time-of-flight VLP16 lidar sensors which actively emit light to measure distances to surfaces.  Unlike the imaging data which represents dense samples of incoming light, the lidar data is a swept sequence of timestamped 3D line segments represented by an origin and termination position. A single lidar segment indicates that during the timestamp, the space traversed by an emitted ray did not intersect an opaque surface. We make a simplifying assumption that most surfaces detected by lidar are stationary like buildings and roads, so we can ignore the timestamp information and assume that empty space is empty throughout the entire capture. This allows us to model lidar rays similar to camera rays, with origin $\origin_\ell$, direction $\dir_\ell$, and termination distances $z_\ell$.


\section{Method}
\label{sec:method}

We define a Urban Radiance Field (URF) with scene-level neural network parameters~$\params$ as well as per-image exposure parameters~$\{\latentExposure_i\}$.   Given the image and lidar data for a scene, we optimize a URF by minimizing the following loss:
\begin{align}
\argmin_{\params,\{\latentExposure_i\}} \:\:
&\underbrace{\loss{rgb}(\params, \{\latentExposure_i\})
+
\lossSky(\params)
}_{\Sec{rgb}} + \underbrace{\loss{depth}(\params) + \loss{sight}(\params)}_{\Sec{lidar}}
\nonumber
\end{align}


\subsection{Photometric-based Losses}
\label{sec:rgb}
%
The photometric loss term is similar to the original NeRF equation \eq{nerfrgb}, but ours also depends on estimated per-image exposure parameters $\{\latentExposure_i\}$: 
\begin{equation}
\loss{rgb}(\params, \{\latentExposure_i\}) = \sum_{i} \expect_{\ray \sim \image_i}
\left[
||\C(\ray ; \latentExposure_i) - \C^\gt_i(\ray) ||_2^2 
\right]
\label{eq:rgb}
\end{equation}
We modify the volume rendering equation in two ways, each described in its corresponding sub-section:
\begin{align}
\C(\ray; \latentExposure_i) = \int_\near^\far
w(t)
\cdot
\!\underbrace{\exposure(\beta_i)}_\text{\Sec{exposure}} \! \cdot \: {\radiance(t)} \, dt + \underbrace{\radiance_\sky(\dir)}_\text{\Sec{sky}}
\label{eq:color}
\end{align}

\subsubsection{Exposure compensation}
\label{sec:exposure}
Images acquired by mapping systems are usually captured with auto white balance and auto exposure which complicates the computation of $\loss{rgb}$ in \eq{nerfrgb}.
Previous work addresses this issue using latent codes, learned separately for each image, that map image-independent scene radiance to an image-dependent radiance~\cite{adop,neuralrgbd,nerf_w}. 
One shortcoming with such an approach is that modeling exposure variations with a per-image latent code is overparameterized as it allows the latent codes to compensate for non-exposure related errors.
Instead, in~\eq{color} we perform an affine mapping of the radiance predicted by the shared network where the affine transformation is a 3x3 matrix decoded from the per-image latent code $\beta_i \in \real^B $:
\begin{equation}
\exposure(\beta_i) \definedAs \exposure(\beta_i; \params) : \real^B \rightarrow \real^{3 \times 3}   
\end{equation}

This mapping models white balance and exposure variations with a more restrictive function, and thus is less likely to cause unwanted entanglement when the scene radiance parameters $\params$ and the exposure mappings $\beta$ are optimized jointly.


\subsubsection{Sky modeling}
\label{sec:sky}
Outdoor scenes contain sky regions where rays never intersect any opaque surfaces, and thus the NeRF model gets a weak supervisory signal in those regions. To address this issue, our rendering model includes a spherical radiance (environment) map represented as a coordinate-based neural network, similar to the radiance map used in GANcraft~\cite{hao2021GANcraft}
\begin{equation}
\radiance_\sky(\dir) \definedAs \radiance_\sky(\dir; \params) : \real^3 \rightarrow \real^3
\end{equation}
to provide a direction-dependent background color for those regions.
To modulate which rays utilize the environment map, we run a pre-trained semantic segmentation model for each image to detect pixels likely to be sky:  $\mathcal{S}_i {=}\mathcal{S}(\image_i)$, where $\mathcal{S}_i(\ray){=}1$ if the ray~$\ray$ goes through a sky pixel in image $i$.  We then use the sky mask to define an additional loss that encourages at all point samples along rays through sky pixels to have zero density:
\begin{equation}
\lossSky(\params) = 
\expect_{\ray \sim \image_i}
\left[
\mathcal{S}_i(\ray) \int_\near^\far w(t)^2 \, dt
\right]
\end{equation}
Note that whenever $\mathcal{S}_i(\ray){=}1$, this will force the $\radiance_\sky$ to explain the pixel for ray $\ray$ in \eq{color}.


\subsection{Lidar losses}
\label{sec:lidar}

Since lidar data is available in our data, we use it to supervise training of the model.  We are given a collection of $L$ lidar samples~$\lidarPoints {=} \{ (\origin_\ell, \dir_\ell, z_\ell)_{\ell=1}^L\}$, each corresponding to a ray $\ray(z){=} \origin_\ell + z \dir_\ell$, and the associated 3D measurement $\lidarPoint_\ell {=} \ray(z_\ell)$.
Inspired by classical works in 3D reconstruction~\cite{kinfu}, we break the losses into two different types: \CIRCLE{1} supervising the expected \textit{depth} value, and \CIRCLE{2} supervising the free space along the \textit{line-of-sight} from the lidar sensor to the observed position.

\paragraph{Expected depth}
We start by supervising the expected depth $\hat z$ from a volumetric rendering process (i.e.~optical depth~\cite{kangle2021dsnerf}) to match the depth of the lidar measurement:
\begin{align}
\loss{depth}(\params) &=
\expect_{\ray \sim \lidarPoints}
\left[
(\hat z - z)^2 
\right]
\ \hat z =\int_\near^\far \!\!\!  w(t) \cdot t\, dt
\label{eq:depth}
\end{align}
%

\paragraph{Line-of-sight priors}
\label{sec:empty}
For points that are observed by a lidar sensor, a reasonable assumption is that a measured point $\lidarPoint$ corresponds to a location on a non-transparent surface, and that atmospheric media \textit{does not} contribute to the color measured w.r.t. a lidar ray $\ray_\ell = \ray(z_\ell)$.
Hence, we expect that the radiance is concentrated at a \textit{single} point along the ray, and therefore that a \textit{single} point is responsible for the observed color.
In other words, with reference to~\eq{nerfcolor}:
\begin{equation}
\C(\ray_\ell) \equiv \radiance(\ray_\ell) \quad\text{iff}\quad w(t)=\delta(t)
\end{equation}
where $\delta(.)$ is the continuous Dirac function.
We can convert this constraint via the penalty method into a loss: 
\begin{align}
\loss{sight}(\params) &=
\expect_{\ray \sim \lidarPoints}
\left[
\int_\near^\far \left(w(t) - \delta(z) \right)^2 \, dt
\right]
\label{eq:sight}
\end{align}
and to make this numerically tractable, we can replace the Dirac with a kernel $\mathcal{K}_\epsilon(x)$ that integrates to one (i.e. a distribution) that has a bounded domain parameterized by~$\epsilon$.
We choose $\mathcal{K}_\epsilon(x){=}\gaussian(0,(\epsilon/3)^2)$, with $\gaussian$ being a truncated Gaussian, and then split the ray integral into three intervals with three corresponding losses:
\begin{align}
\loss{sight}(\params) &= \underbrace{\loss{empty}(\params)}_{t \in [\near, z-\epsilon]} + \underbrace{\loss{near}(\params)}_{t \in [z-\epsilon, z+\epsilon]} + \underbrace{\loss{dist}(\params)}_{t \in [z+\epsilon,\far]}
\end{align}
The second term in the breakdown above will be:
\begin{align}
\lossNear(\params) &=
\expect_{\ray \sim \lidarPoints}
\left[
\int_{z-\epsilon}^{z+\epsilon} \left(w(t) - \mathcal{K}_\epsilon(t-z) \right)^2 \, dt
\right]
\label{eq:near}
\end{align}
which encourages the representation to increase volumetric density in the neighborhood of $\lidarPoint$, thereby allowing training to converge more quickly.
Note that, as~$\mathcal{K}_\epsilon(x)$ has bounded support in~$[z-\epsilon, z+\epsilon]$, the first term can be simplified to:
\begin{align}
\lossEmpty(\params) &=
\expect_{\ray \sim \lidarPoints}
\left[
\int_\near^{z-\epsilon} w(t)^2 \, dt
\right]
\label{eq:empty}
\end{align}
which requires that the portion of space between the ray origin and the lidar point~$\lidarPoint$ (i.e. the line-of-sight) does not contain any 3D surface.
This line of sight information has been a key ingredient in ``volume carving'' techniques~\cite{volrange, kinfu, visualhull}.
The last term has a similar form:
\begin{align}
\loss{dist}(\params) &=
\expect_{\ray \sim \lidarPoints}
\left[
\int_{z+\epsilon}^\far w(t)^2 \, dt
\right],
\label{eq:dist}
\end{align}
however, because this term's only purpose is to ensure that $w(t)$ sums to one, and because NeRF's volume rendering equation only requires that $w(t)$ sums to no more than one, this term can be safely dropped during training.
Note that the choice of a smooth kernel $\mathcal{K}_\epsilon(x)$ is critical, as it guarantees continuity across the transition between losses at~$z-\epsilon$.
Finally, selecting a suitable $\epsilon$ plays an important role in the reconstruction accuracy.
We discovered that employing a small $\epsilon$ hinders performance, especially in the early training phases, and note that a similar behavior has also been observed in somewhat related methods that anneal the bandwidth of importance sampling over time~\cite{Oechsle2021ICCV}.
In our network, we adopt an exponential decay strategy for $\epsilon$, and ablate this decision in~\supplementary.
\section{Experimental Evaluation}
\label{sec:Evaluation}

We ran a series of experiments to evaluate our model and to test whether the proposed new ideas enable more accurate renderings of novel views and improve the quality of 3D geometric reconstructions.


\subsection{Evaluation Protocol}

\paragraph{The Street View dataset}
We collected 10 scenes from cities around the globe covering six continents. Each scene corresponds to a trekker capture of approximately 20 panoramas (each containing 7 images) and 6 million lidar points on average. Each scene covers hundreds of square meters and represents a different urban environment. For the quantitative analysis we report average metrics over the scenes \textit{Taipei, Zurich, New York, Rome}.
We split this overall dataset in two different ways in a training and a test set, giving rise to the two experimental settings below.

\paragraph{Setting 1: Held-out Viewpoints}
We split each scene into train and test based on the camera locations. We randomly select $20\%$ simultaneous image captures from our trekker rig and use them as test views.
As lidar sensors operate in a continuous fashion, we select all lidar rays whose origins are close to a test camera's location as the lidar test set.

\paragraph{Setting 2: Held-out Buildings}
We also want to evaluate how well our model reconstructs entire 3D surfaces for which we do not have any lidar.
To simulate this, we manually select a building and remove all lidar rays terminating on its surface; these removed rays form the test set. We use the remaining lidar rays and all images as the training set.

\subsection{Baseline Methods}

We perform comparisons with the following baselines
\footnote{We experimented also with NerfingMVS~\cite{Wei21iccv_NerfingMVS}, but the results were much lower than the other methods.}.
For each of them, we adjust the model parameters (number of rays, samples, etc) to be comparable: 
\begin{itemize}[leftmargin=*]
\setlength\itemsep{-.3em}
\item \textbf{NeRF~\cite{Mildenhall20eccv_nerf}} -- We use the JAX~\cite{jax2018github} version, which is a superior re-implementation~\cite{jaxnerf2020github} of the original NeRF paper. This method operates on images only.
\item \textbf{Mip-NeRF~\cite{barron2021mipnerf}} -- an extension of NeRF that uses integrated positional encoding. It also operates only on images and is the method that we build upon.
\item \textbf{DS-NeRF~\cite{kangle2021dsnerf}} -- this paper used 3D keypoints from an SfM reconstruction to supervise the NeRF density; here we adjust it to use lidar points (the same we also use).
\item \textbf{NeRF-W~\cite{MartinBrualla21cvpr_nerfw}} -- NeRF in the Wild has shown impressive results on outdoor scenes and can handle images with different exposures. It operates on images only.
\end{itemize}

\begin{table}
\begin{center}
\begin{tabular}{@{\,}llccc@{\,}} 
 & Lidar& PSNR $\uparrow$ &  SSIM $\uparrow$ & LPIPS $\downarrow$ \\ 
\hline
NeRF\cite{Mildenhall20eccv_nerf} & \untickbox& 14.791 & 0.477 & 0.569  \\
NeRF-W\cite{nerf_w} & \untickbox&17.156 & 0.455 & 0.620 \\
Mip-NeRF\cite{barron2021mipnerf} & \untickbox&16.987 & 0.516 & 0.458         \\
DS-NeRF\cite{kangle2021dsnerf} &\tickbox& 15.178  & 0.500 & 0.537  \\
\hline
Ours w/o lidar &\untickbox& 19.638 &0.541  & 0.447 \\
\textbf{Ours} &\tickbox&   \textbf{20.421}&  \textbf{0.563}  &   \textbf{0.409} \\

\end{tabular}
\end{center}
\vspace{\vspacebeforecaption}
\caption{\textbf{Novel view synthesis} --
We report standard image rendering metrics on test views of selected scenes.
}
\label{tab:\currfilebase}
\end{table}

\begin{figure*}[t]
\begin{center}
\includegraphics[width=\linewidth]{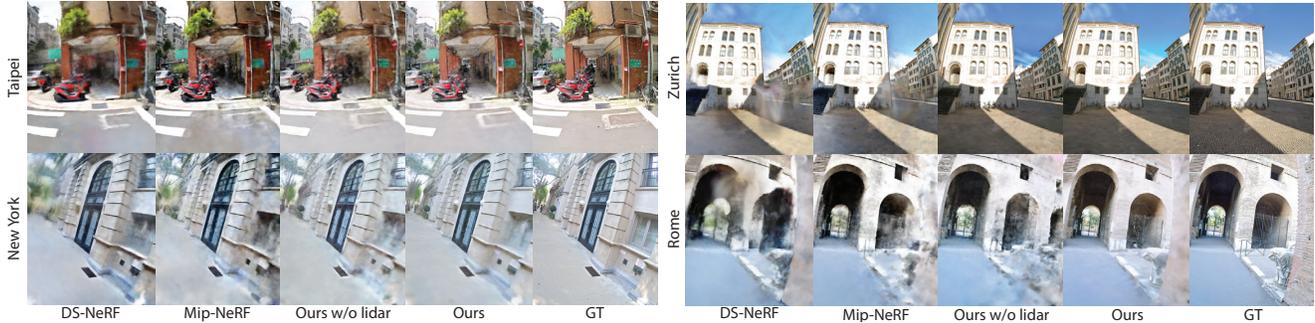}
\end{center}
\vspace{\vspacebeforecaption}
\caption{
\textbf{Qualitative novel view synthesis -- }
We visualize the output of our model against the ground truth and different methods. Our full model is able to generate more accurate renderings that do not suffer from exposure artifacts and floating elements.
}
\label{fig:\currfilebase}
\vspace{\vspaceaftercaption}
\end{figure*}
\subsection{Novel View Synthesis Results}
We first consider novel view synthesis by training and then rendering novel views in the \textit{Held-out Viewpoints} setting (Tab.~\ref{tab:novel_view_synthesis}).
We evaluate the rendered test views using three standard metrics: PSNR, SSIM, and LPIPS~\cite{zhang2018perceptual}.
Similar to the protocol of \cite{MartinBrualla21cvpr_nerfw}, we evaluate on the right part of the image, as the left is used for test-time optimization of the exposure latent codes for our method and NeRF-W. 

Starting from the base model Mip-NeRF, we achieve significant improvements by adding our exposure and sky modeling (second-last row).
Additionally, including lidar information improves the renderings even further (last row).
The methods we compare against are challenged by Street View data due to its sparsity in terms of viewpoints and exposure variations between images. We outperform all of them, including NeRF-W which was designed for outdoor scenes, and DS-NeRF, which exploits lidar.

In Fig.~\ref{fig:rgb_comparison}, we show rendered results on test views from various models. Mip-NeRF suffers from its inability to handle exposure variations, resulting in floating artifacts that attempt to explain the differences in exposure between training views.
Next we find that our full model shows significantly sharper images when using lidar due to geometrically accurate surface placements and suppression of erroneous, semi-transparent density ``floaters''.
The improvement is more visible in the distant areas like the arcade in Rome and the covered sidewalk in Taipei.


\subsection{3D Reconstruction Results}
\label{sec:3d-results}

Next we evaluate the quality of the recovered 3D scene structure, both in terms of depth estimates and point clouds.

\begin{figure*}[t]
\begin{center}
\includegraphics[width=\linewidth]{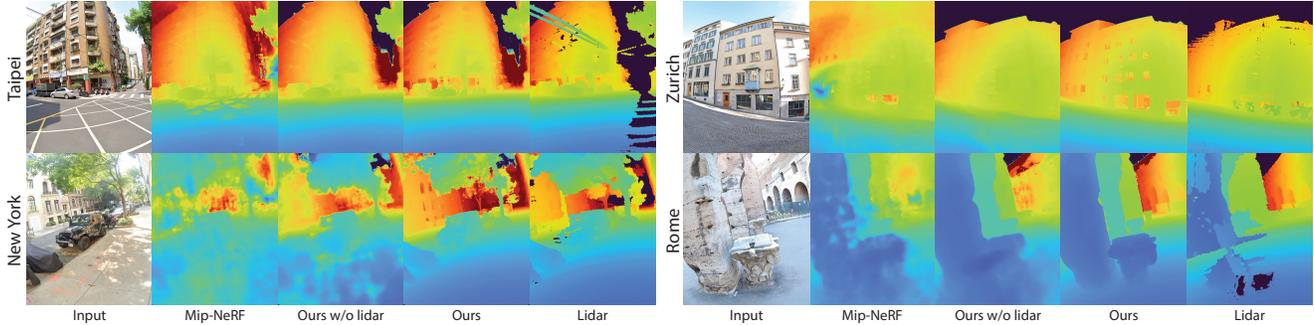}
\end{center}
\vspace{\vspacebeforecaption}
\vspace{-.5em}
\caption{
\textbf{Qualitative depthmaps -- }
We visualize the expected depth for our model against other methods and variations. Our full model is able to estimate precisely the extent of the scene, including thin structures such as tree trunks, window frames, etc. 
}
\label{fig:\currfilebase}
\vspace{\vspaceaftercaption}
\end{figure*}
\paragraph{Depth estimates}
As shown in Fig.~\ref{fig:depthmaps}, our full model is able to use the sparse lidar supervision to reconstruct significantly finer depth detail compared to using only the dense pixel supervision. In particular, we reconstruct crisp depth values even for some surfaces that are difficult to capture, like cars and window frames. Note that the lidar depthmaps are estimated using depth-aware splatting and the missing regions are due to the lidar scanning pattern.
%
For quantitative evaluation, we use the set of test lidar rays' origin $\origin_\ell$ and direction $\dir_\ell$ to cast rays and ask the model to estimate the expected termination distance $\hat{z}$ by sampling its volumetric density function and accumulating its transmittance, similar to Eq.~\eqref{eq:depth}. We compare this model estimate to the ground truth termination distance $z$ and report the average error in meters.
We also report accuracy as the number of test rays estimated within~$0.1$ meters of their ground truth~(Acc@0.1m).
Looking at the results in Tab. \ref{tab:point_set_evaluation}, we find that our model outperforms all baselines on both metrics (by $\sim\!3\times$). 
\begin{table*}
\begin{center}
\begin{tabular}{ llcccc|cccc } 
& & \multicolumn{4}{c}{Held-out Viewpoints} &  \multicolumn{4}{c}{Held-out Building} \\
  & Lidar &  Avg Error (m) $\downarrow$  &  Acc$\uparrow$  & CD$\downarrow$ & F$\uparrow$ &   Avg Error (m) $\downarrow$  &  Acc$\uparrow$ & CD$\downarrow$   & F$\uparrow$ \\ 
  \hline
NeRF\cite{Mildenhall20eccv_nerf}     &\untickbox &   1.582 & 0.264 & 3.045 & 0.528 &    1.423 & 0.274 & 2.857 & 0.535 \\
NeRF-W\cite{nerf_w}       &\untickbox &   3.663 & 0.144 & 6.165 & 0.372 &    1.348 & 0.207 & 4.054 & 0.552 \\
Mip-NeRF\cite{barron2021mipnerf}    &\untickbox &   1.596 & 0.133 & 2.812 & 0.363 &    1.417 & 0.132 & 2.508 & 0.427 \\
DS-NeRF\cite{kangle2021dsnerf}   &\tickbox &   1.502 & 0.259 & 2.571 & 0.526 &    1.367 & 0.294 & 2.720 & 0.558 \\
  \hline
Ours                                &\tickbox &   \textbf{0.463} & \textbf{0.742} & \textbf{0.272} & \textbf{0.880} &    \textbf{0.770} & \textbf{0.363} & \textbf{2.312} & \textbf{0.687} \\
\end{tabular}
\end{center}
\vspace{\vspacebeforeTABLEcaption}
\caption{
\textbf{Reconstruction evaluation} --
We compare various NeRF based approaches in two experimental settings (\textit{Held-out Viewpoints}, \textit{Held-out Buildings}). We report two depth estimation metrics (Average Error and Accuracy) and two point cloud metrics (Chamfer Distance CD, and F-score). See main text for details.
}
\label{tab:\currfilebase}
\end{table*}

\paragraph{Point clouds}
We generate 3D point clouds directly from the ray parameters and depth estimates. Given a ray origin $\origin_\ell$, direction $\dir_\ell$ and ground truth termination distance $z$, the corresponding 3D point is $\lidarPoint_\ell=\origin_\ell +z\dir_\ell$. By iterating over all the test lidar rays we estimate the ground truth point cloud of the scene. We do the same for the estimated depth $\hat{z}$, resulting in the predicted point cloud. We compare the two clouds using Chamfer Distance and F-score (threshold $=$ 0.1 meters).
As Tab. \ref{tab:point_set_evaluation} shows, our model has the best performance across all metrics in both held-out settings. The difference is large even over DS-NeRF, that uses lidar, indicating that both our exposure/sky modeling and our combination of losses are important to achieve high accuracy.


\begin{figure}[t]
\begin{center}
\includegraphics[width=\linewidth]{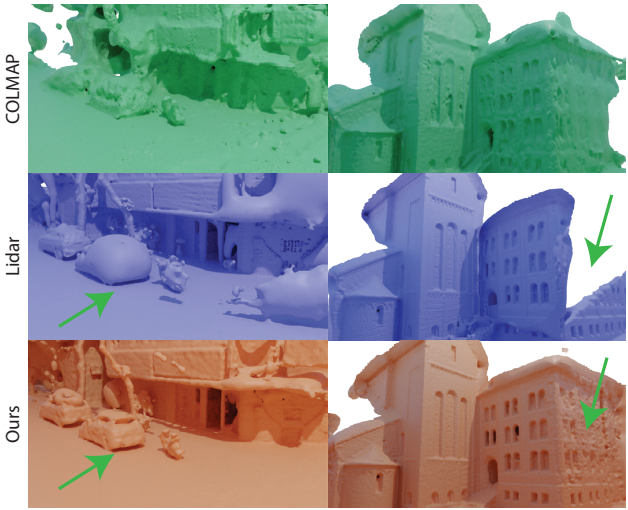}
\end{center}
\vspace{\vspacebeforecaption}
\caption{
\textbf{Surface Reconstruction -- }
We show the surface reconstruction returned by different approaches. Our method is able to provide dense and accurate depth estimates, which in turn allow detailed mesh reconstructions.
}
\label{fig:\currfilebase}
\vspace{\vspaceaftercaption}
\end{figure}
\begin{figure}[t]
\begin{center}
\includegraphics[width=\linewidth]{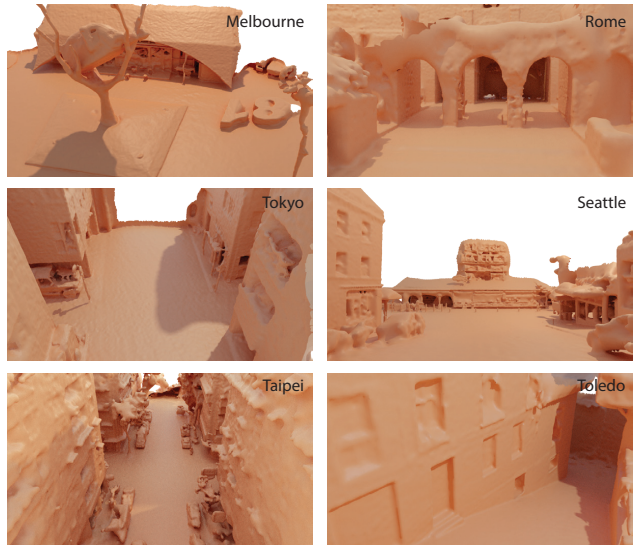}
\end{center}
\vspace{\vspacebeforecaption}
\caption{
\textbf{Additional Reconstructions -- }
Visualization of extracted meshes for several large scale urban scenes.
}
\label{fig:\currfilebase}
\vspace{\vspaceaftercaption}
\end{figure}
\paragraph{Mesh reconstruction}
Here we use our full model to generate dense point clouds by casting one ray for each pixel in each training camera, and estimating depth for it as described above.
For comparison, we obtain a point cloud with COLMAP~\cite{schoenberger2016mvs} estimated by running MVS using the camera parameters provided with the dataset. 
We also compare to the point cloud defined by the lidar points (the training points  that we also feed to our method).
For each method, we then reconstruct 3D meshes using Poisson Surface Reconstruction~\cite{KazhdanSGP06}.

Fig.~\ref{fig:reconstruction_comparison} shows the meshes derived
from our point clouds, from COLMAP and from lidar.
Our method is able to estimate accurately the underlying geometry, whereas COLMAP loses fine details and lidar produces artifacts due to limited sampling resolution. Our method also provides denser coverage than the raw lidar, since we also use images. These provide higher resolution observations in some regions (e.g., the cars in the image on the left) and broader coverage (e.g., the missing region of the building on the right), as the scanning pattern of the lidar is far narrower than the image panoramas.
Fig.~\ref{fig:recon} shows more mesh reconstructions from our method. 
We can accurately reconstruct fine details in areas hundreds of square meters large.


\subsection{Ablation Studies}
\label{sec:ablations}

\begin{table}
\begin{center}
\begin{tabular}{ p{0.2cm}  p{0.2cm} p{0.2cm} p{0.2cm} p{0.2cm}  cccc } 
 \rotatebox{90}{$\exposure$} & \rotatebox{90}{$\lossSky$} & \rotatebox{90}{$\lossDepth$} &  \rotatebox{90}{$\lossEmpty$} &  \rotatebox{90}{$\lossNear$} & Avg$\downarrow$  &  Acc$\uparrow$ & CD$\downarrow$   & F$\uparrow$\\
\hline
\untickbox & \untickbox & \untickbox & \untickbox & \untickbox & 1.596 &	0.133& 2.812&		0.363 \\
\tickbox & \untickbox & \untickbox & \untickbox & \untickbox   & 1.226 &	0.184&   1.771&		0.424 \\
\tickbox & \tickbox & \untickbox & \untickbox & \untickbox& 1.109&	0.233&     1.190&		0.471 \\
\hdashline

\tickbox & \tickbox & \tickbox & \untickbox & \untickbox& 0.811&	0.284&     0.782&		0.545 \\
\tickbox & \tickbox & \tickbox & \tickbox & \untickbox& 1.136&	0.093&       0.536&		0.306 \\
\tickbox & \tickbox & \untickbox & \tickbox & \tickbox& 0.633&	0.736&       0.726&		0.878 \\
\tickbox & \tickbox & \tickbox & \tickbox & \tickbox& \textbf{0.463}&	\textbf{0.742}&         \textbf{0.272}&		\textbf{0.880} \\
\end{tabular}
\end{center}
\vspace{\vspacebeforeTABLEcaption}
\caption{
\textbf{Ablation study} --
We investigate the effect of the individual components of our model, from the image-based elements (top part) to the lidar losses (bottom part). The first row corresponds to the baseline Mip-NeRF~\cite{barron2021mipnerf} model.
}
\label{tab:\currfilebase}
\end{table}
\paragraph{Effects of individual components}
Tab. \ref{tab:ablation} studies the effect of each model component.
As we add exposure compensation and sky modeling we see consistent improvements in all 3D reconstruction metrics (second and third rows).
When incorporating lidar, the best performing setup is when using all proposed losses at the same time (Sec. \ref{sec:lidar}).
The strongest contribution comes from the near-surface loss, as when its not activated the performance drops considerably.
The empty-space loss, which is primarily used to suppress floating semi-transparent density (``floaters''), harms performance when used without the near-surface loss.
This indicates that our decaying margin strategy is suitable way to gather the benefits of both losses.
We further analyze this behaviour in the \supplementary.


\begin{figure}[h]
\begin{center}
\includegraphics[width=\linewidth]{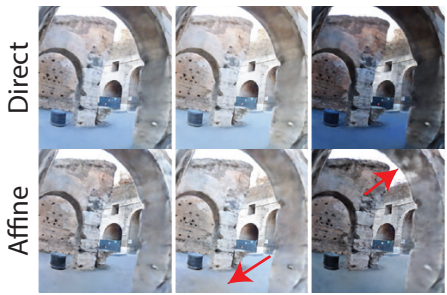}
\end{center}
\vspace{\vspacebeforecaption}
\caption{
\textbf{Exposure modifications -- }
We visualize the effect of changing exposure codes using our affine color model and a direct approach. The affine model performs a global color transformation, enforcing the disentanglement between exposure codes and scene radiance.
}
\label{fig:\currfilebase}
\vspace{\vspaceaftercaption}
\end{figure}
\paragraph{Effect of exposure handling}
Finally, we investigate the effect of our affine color transformation.
An alternative is to provide the exposure code $\beta_i$ directly as an input to the network, as done in NeRF-W~\cite{MartinBrualla21cvpr_nerfw} and \cite{neuralrgbd}. 
However, in this way the exposure code can explain arbitrary appearance elements, and not necessarily those due to exposure/white balance.
Fig.~\ref{fig:exposure} illustrates the appearance changes when modifying the latent codes on the \textit{Rome} scene.
For our affine model this translate into rendering the same structure with different color tones, while the direct approach generates visible artifacts. 
These affect also 3D reconstruction performance: our affine model results in F-score of $0.47$ vs $0.36$ for the direct approach,
These artifacts can be reduced by limiting the power of the latent codes to affect rendering, \eg by providing $\beta_i$ to later layer of the network, or reducing their dimensionality (similar to observations in~\cite{kaizhang2020_nerfplusplus} regarding the viewing direction $\mathbf{d}$). See the \supplementary for more details.


\begin{figure}[t]
\begin{center}
\includegraphics[width=\linewidth]{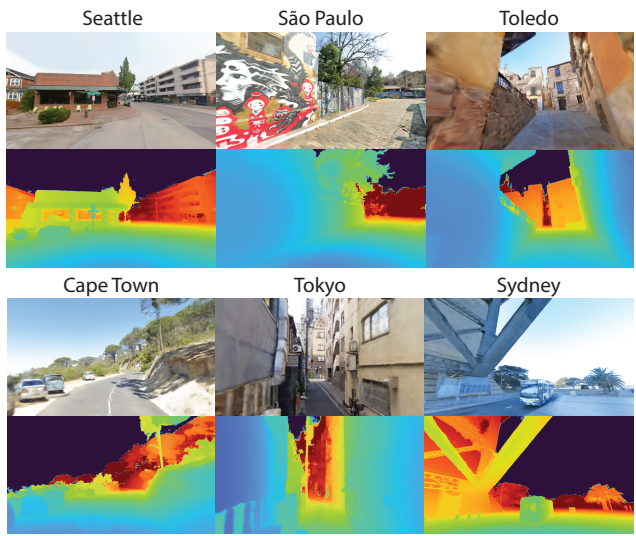}
\end{center}
\vspace{\vspacebeforecaption}
\caption{
\textbf{Novel views -- }
After training our model for a particular scene, we are able to estimate novel views for a new camera with different intrinsics and trajectory path than those used for training. 
}
\label{fig:\currfilebase}
\vspace{\vspaceaftercaption}
\end{figure}
\begin{figure}[h]
\begin{center}
\includegraphics[width=\linewidth]{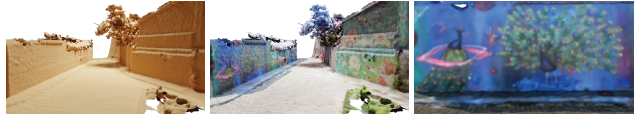}
\end{center}
\vspace{\vspacebeforecaption}
\caption{
\textbf{Colored mesh -- }
Using our model we can also accurately reconstruct a colored scene mesh.
}
\label{fig:\currfilebase}
\vspace{\vspaceaftercaption}
\vspace{-5pt}
\end{figure}
\subsection{Qualitative results}
\vspace{-5pt}
A benefits of having precise 3D reconstruction as part of a NeRF-based model is that it enables great flexibility in placing a virtual camera for novel view synthesis.
In Fig.~\ref{fig:novel_views} we visualize rendered images and depth maps by our model from camera positions substantially different from those along the trekkers acquisition path (on average 1.7 meters away from the closest training camera).

In Sec. \ref{sec:3d-results} we showed how we can estimate accurately the scene geometry as meshes.
In Fig. \ref{fig:colored_mesh}, we illustrate how we can use our model to query the colors of the mesh vertices, resulting in precise textured meshes that are compatible with traditional rendering software. Note that the exposure is already compensated, while traditional pipelines need to perform screened Poisson image integration to compensate for the color variation~\cite{dessein2014seamless},


\vspace{-5pt}
\subsection{Limitations}
\vspace{-5pt}
Our system has limitations. 
First, it assumes that good camera parameters are given through an SfM pipeline, but they are sometimes noisy in practice --- a joint optimization of the camera parameters along with the network parameters may improve the reconstructions for those cases~\cite{Lin21iccv_BARF,Wang21arxiv_NeRFminusminus}.
Second, it has been demonstrated only for snippets of data cut out of longer scanning sequences --- models learned from multiple snippets would have to be \textit{stitched} together to produce a coherent model of the spatially continuous world.
These limitations, plus failure cases and potential negative societal impacts, are discussed further in the supplemental material.
\vspace{-5pt}

\section{Conclusion}
\label{sec:conclusion}
\vspace{-5pt}

We present a system for 3D reconstruction and novel view synthesis from data captured by mobile scanning platforms in urban environments.
Our approach extends recent work on Neural Radiance Fields with new ideas to leverage asynchronously captured lidar data, to account for differences in exposures between captured images, and to leverage predicted image segmentations to supervise the density of rays pointing towards the sky.
Experimental results on Street View data demonstrate that each of these three ideas significantly improves performance on its own, and that they combine to produce better synthesized novel views (+19\% PSNR over \cite{MartinBrualla21cvpr_nerfw}) and 3D surface reconstructions (+0.35 F-score over \cite{kangle2021dsnerf}).
We hope this paper inspires future work to take further steps towards deploying coordinate-based neural networks in outdoor mapping applications.

\clearpage
{
    \small
    \bibliographystyle{ieee_fullname}
    \bibliography{macros,main,nerf}
}
\clearpage

\twocolumn[
\centering
\Large
\textbf{URF: Urban Radiance Fields} \\
\vspace{0.5em}Supplementary Material \\
\vspace{1.0em}
] 
\appendix

The following supplemental material contains additional implementation details, ablation studies and qualitative results.

\section{Additional Implementation Details}
\label{sec:implementation}

\subsection{Network architecture}
\label{sec:architecture}

\begin{figure}[h]
\begin{center}
\includegraphics[width=\linewidth]{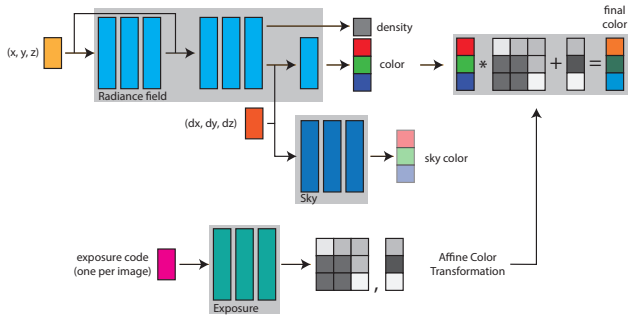}
\end{center}
\vspace{\vspacebeforecaption}
\caption{
\textbf{Network architecture}
}
\label{fig:\currfilebase}
\vspace{\vspaceaftercaption}
\end{figure}
Our network architecture is illustrated in Fig.~\ref{fig:network_architecture} and has three components.
The first component is the neural radiance field network, which is similar to the original NeRF~\cite{Mildenhall20eccv_nerf}. It consists of a series of fully connected layers of width 256 that take as input the 3D location of a point $x, y, z$ and the viewing direction $dx, dy, dz$ and output the RGB color and the density at that point.
The second component is the sky network, which takes as an input the direction $dx, dy, dz$ of a ray pointing at a sky point, and outputs its color.
Finally, the third component is an exposure compensation network that takes as an input an exposure latent code and estimates the affine transformation to be applied to the color values output by the radiance field network. There is a different affine transformation per image. This compensates for the different exposures across input images.
All three network are trained jointly so that the final colors output by the model will match the pixel colors in the input images.

\subsection{Training Protocol}
\label{sec:training}

We train a separate network for each baseline model (Sec. 5.2) and each variant of our model, applies to each scene.
Every network is trained with the same protocol, detailed here.
We use a TPU v2 architecture with 128 cores~\cite{tpu} using Tensorflow 2~\cite{tensorflow2015-whitepaper}.
We used the Adam optimizer~\cite{Kingma15Adam} with a learning rate scheduler that included two stages. The warm up stage lasts 50 epochs, with the learning rate starting at 0.0005 and growing linearly until 0.005.
After warm up, the main stage lasts 500 epochs, with the learning rate starting at 0.005 and then decaying exponentially with exponent 0.98.
The ray batch size was set to 2048 per core and the total training time was about one day per network.

For ray sampling, we use a stratified strategy where the intervals are evenly spaced in log scale, and we sample 1024 samples per ray. We did not perform hierarchical sampling. Each batch contains rays randomly sampled from all images (and similarly for the lidar points) 

The 3D location of a point is described using integrated positional encoding~\cite{barron2021mipnerf} with $L=10$ frequencies. For the viewing direction we use the original positional encoding representation with 4 frequencies.


\section{Additional Ablation Studies}
\label{sec:ablations}

\subsection{Effect of margin $\epsilon$}
\label{sec:margin}

\begin{table}[h]
\begin{center}
\begin{tabular}{ lcccc } 
  &  Avg Error$\downarrow$ \ & CD$\downarrow$ &  Acc$\uparrow$  & F$\uparrow$ \\ 
  \hline
Fixed       &1.007&2.195&0.814&0.871\\
Stepwise     &0.776&1.818&0.849&0.905\\
Linear       &0.238&0.508&0.903&0.961\\
Exponential  &0.249&0.863&0.901&0.966\\

\end{tabular}
\end{center}
\vspace{\vspacebeforeTABLEcaption}
\caption{
\textbf{Margin decay ($\epsilon$)} --
We evaluate different decay strategies for the margin $\epsilon$ during training in the Rome scene. The margin controls the contribution of the lidar losses $\loss{near}$ and $\loss{empty}$. Having a fixed margin results in lower performance, while gradually decreasing it performs the best.
}
\label{tab:\currfilebase}
\end{table}

As we observed in Sec. 5.5 of the main paper, the empty-space loss can actually decrease 3D reconstruction performance as it introduces a strong preference for empty space.
Using the near-surface loss, which is complementary to empty-space by construction, alleviates this effect.
In Tab. \ref{tab:margin} we vary the margin $\epsilon$ in Eq.~(16) and Eq.~(17) during training using different strategies:
Fixed: keep a constant margin throughout training (as in~\cite{neuralrgbd});
Stepwise: start with a large margin (thus only the near-surface loss is activated) and after $N=50$ epochs the margin suddenly becomes small;
Linear/exponential: gradually reduce the margin from large to small with a linear or an exponential schedule.
For all methods, the smallest value $\epsilon$ was set to $20$cm.
The linear and exponential strategies perform similarly and much better than the fixed and stepwise ones, indicating that the empty-space loss is best applied after the training process manages to infer a good initial version of the scene structure.


\subsection{Effect of exposure handling}
\label{sec:exposure}
\begin{table}[ht]
\begin{center}
\begin{tabular}{ llllll } 
 &$D$&  Avg Error$\downarrow$ \ & CD$\downarrow$ &  Acc$\uparrow$  & F$\uparrow$ \\ 
\toprule
Direct& \multirow{2}{*}{48}  &1.071	&1.28	&0.159	&0.362\\
Affine       & &0.98	&1.007	&0.253	&0.47\\
\midrule
Direct     &\multirow{2}{*}{4} &1.049	&2.062	&0.247	&0.477\\
Affine       & &0.885	&1.564	&0.262	&0.524\\
\end{tabular}
\end{center}
\vspace{\vspacebeforeTABLEcaption}
\caption{
\textbf{Exposure handling} -- We compare our affine transformation model with the direct input of the exposure code to the network. Using an explicit color transformation for the different exposures results in better reconstruction. 
}
\vspace{-10pt}
\label{tab:\currfilebase}
\end{table}

In Tab.~\ref{tab:exposure} we expand the ablation experiment in Sec.~5.5 of the main paper, comparing our affine transformation model versus directly providing the exposure latent code to the network for the Rome scene.
We experiment with two different dimensions for the latent codes, $D=48$ as in NeRF-W~\cite{MartinBrualla21cvpr_nerfw} and a much smaller one $D=4$.
As Tab.~\ref{tab:exposure} shows, our affine transformation approach performs better in all 3D reconstruction metrics, for both values of $D$.
We also observe that both the affine and the direct approach perform better when the exposure latent code has smaller dimensionality, thus reduced capacity.
For the direct approach this is in accordance to the observations in NeRF++~\cite{kaizhang2020_nerfplusplus} about the viewing direction: implicit regularization (limiting the capacity) of the latent code can increase the performance.
For the affine case, this indicates that there exist a compact latent space that can describe the color transformations appearing in the dataset and it is easier to learn.


\begin{figure}[t]
\begin{center}
\includegraphics[width=\linewidth]{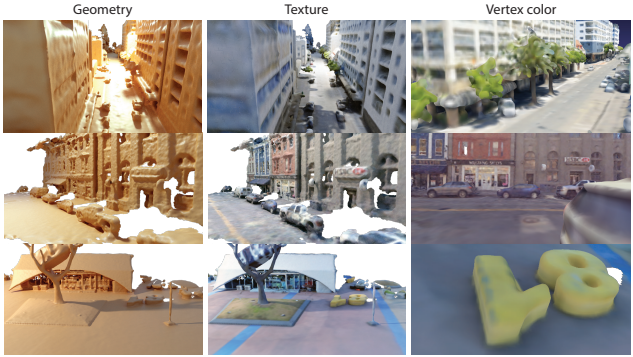}
\end{center}
\vspace{\vspacebeforecaption}
\caption{
\textbf{Rendering colored meshes -- }
After extracting a color mesh using our model, we can render its geometry and texture in a 3D environment, or render the vertex colors in real time on a browser.
}
\label{fig:\currfilebase}
\vspace{\vspaceaftercaption}
\end{figure}
\begin{figure*}[t]
\begin{center}
\includegraphics[width=\linewidth]{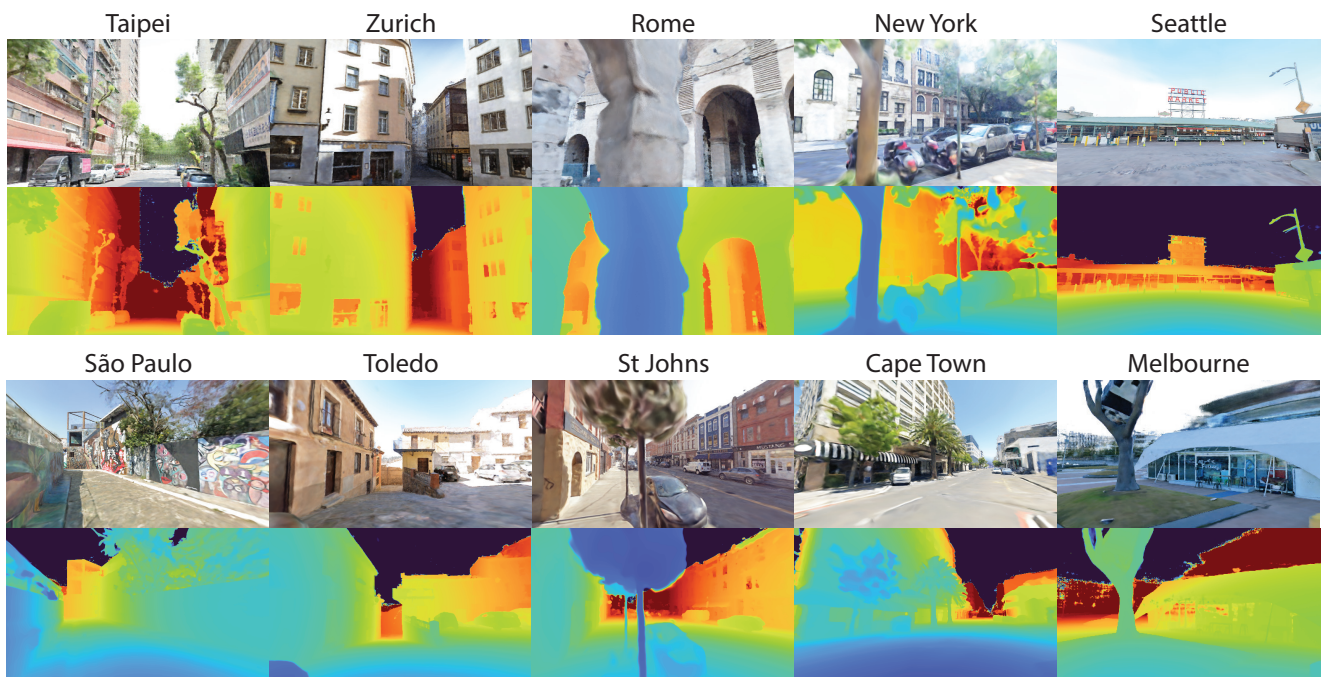}
\end{center}
\vspace{\vspacebeforecaption}
\caption{
\textbf{Novel views}
}
\label{fig:\currfilebase}
\vspace{\vspaceaftercaption}
\end{figure*}
\section{Additional Qualitative Results}
\label{sec:additional_nesults}
In Fig.~\ref{fig:colored_mesh_more} we show more visualizations of extracted colored meshes for different scenes. The color of every vertex is estimated by querying the radiance field network in that particular location. This representation can be used in common 3D editing software such as Blender~\cite{blender} (first and second column in Fig.~\ref{fig:colored_mesh_more}) and it allows for real time rendering on the browser, \eg using ThreeJS~\cite{threejs} (third column in Fig.~\ref{fig:colored_mesh_more}). Note that this way of rendering is different than the volumetric rendering in NeRF models, which is continuous and incorporates implicitly the view dependent appearance changes.
Finally, we present additional results for novel view synthesis in Fig.~\ref{fig:novel_views_more}.


\vspace{-5pt}

\end{document}